# Towards an Evaluation Framework for Explainable Artificial Intelligence Systems for Health and Well-being

Esperança Amengual-Alcover[1][a], Antoni Jaume-i-Capó[1][b], Miquel Miró-Nicolau[1][c], Gabriel Moyà-Alcover[1][d] and Antonia Paniza-Fullana[2][e]

[1] Department of Mathematics and Computer Science, University of the Balearic Islands, Ctra. de Valldemossa, Km. 7.5, 07122 - Palma de Mallorca, Spain.

[2] Department of Private Law, University of the Balearic Islands, Ctra. de Valldemossa, Km. 7.5, 07122 - Palma de Mallorca, Spain.

{eamengual, antoni.jaume, miquel.miro, gabriel.moya, antonia.paniza}@uib.es

Abstract: The integration of Artificial Intelligence in the development of computer systems presents a new challenge: make intelligent systems explainable to humans. This is especially vital in the field of health and well-being, where transparency in decision support systems enables healthcare professionals to understand and trust automated decisions and predictions. To address this need, tools are required to guide the development of explainable AI systems. In this paper, we introduce an evaluation framework designed to support the development of explainable AI systems for health and well-being. Additionally, we present a case study that illustrates the application of the framework in practice. We believe that our framework can serve as a valuable tool not only for developing explainable AI systems in healthcare but also for any AI system that has a significant impact on individuals.

## 1 INTRODUCTION

The third wave of Artificial Intelligence (AI) systems is characterized by two key aspects: (1) technological advancements and diverse applications and (2) a human-centred approach (Xu 2019). While AI is achieving impressive results, these outcomes are often challenging for human users to interpret. To trust the behaviour of intelligent systems, especially in health and well-being, they need to clearly communicate the rationale behind their decisions and actions. Explainable Artificial Intelligence (XAI) aims to meet this need by prioritizing transparency, enabling AI systems to describe the reasoning behind their decisions and predictions.

A growing interest in XAI has been reflected in several scientific events (Adadi and Berrada 2018; Alonso, Castiello, and Mencar 2018; Anjomshoae et al. 2019; Biran and Cotton 2017; Došilović, Brčić, and Hlupić 2018) and in the relevant increase of recent reviews about the topic (Abdul et al. 2018; Alonso, Castiello, and Mencar 2018; Anjomshoae et al. 2019; Chakraborti et al. 2017; Došilović, Brčić, and Hlupić 2018; Gilpin et al. 2018; Murdoch et al. 2019), particularly in the health and well-being areas (Mohseni, Zarei, and Ragan 2021; Tjoa and Guan 2021) XAI is emerging as a new discipline in need of standardized practices. The diverse goals, design strategies, and evaluation techniques used in XAI have resulted in a range of approaches for creating explainable systems (Mohseni, Zarei, and Ragan 2021). Murdoch et al. (Murdoch et al. 2019) propose a broad categorization of XAI methods into model-based and post-hoc techniques. However, as discussed in (Mohseni, Zarei, and Ragan 2021), achieving effective XAI design requires an integrated

---

[a] https://orcid.org/0000-0002-0699-6684
[b] https://orcid.org/0000-0003-3312-5347
[c] https://orcid.org/0000-0002-4092-6583
[d] https://orcid.org/0000-0002-3412-5499
[e] https://orcid.org/0000-0002-1302-9713

approach that considers the dependencies between design goals and evaluation methods.

In this work, we present an evaluation framework that aims to guide the design of explainable AI systems for health and well-being, with an emphasis on legal and ethical issues. We illustrate our approach through a case study on medical image analysis, an area where we have previous experience. In healthcare, decision-making based solely on unexplainable predictions is insufficient to meet ethical and legal standards. Explainability helps to rationalize AI driven diagnoses, treatment plans, and disease predictions, enhancing understanding for both professionals and patients. Indeed, explainability is recognized as an essential ethical principle for AI systems, ensuring their transparency for end-users (Alcarazo 2022). This principle aligns with the European Parliament's "Report on Artificial Intelligence in a Digital Age" (VOSS, n.d.), which emphasizes transparency and explainability as foundational.

Our goal is to ensure that intelligent systems for medical image analysis adhere to these ethical standards and relevant regulations. Given the significant impact of AI-driven decisions on individuals, it is crucial to protect personal data and inform patients about how these systems are used. Furthermore, medical professionals must be able to comprehend the reasoning behind an AI system´s conclusions to understand the logic guiding its predictions and recommendations.

## 2 PREVIOUS WORK

In (Mohseni, Zarei, and Ragan 2021), a generic framework for designing XAI systems is presented, offering multidisciplinary teams a high-level guideline for developing domain-specific XAI solutions. According to its authors, the framework's flexibility makes it broadly applicable, enabling customization to address specific needs across various fields. In our work, we build upon this framework to integrate design guidelines specifically tailored to meet the requirements of health and well-being applications. Figure 1 provides an overview of the original framework, which serves as the basis for our domain-focused extension. The layered structure links core design goals and evaluation priorities across different research communities, promoting multidisciplinary progress in the field of XAI systems. This structure supports the design steps, starting with the most external level (XAI System Objectives), then considering the needs of end users at the intermediate level (Explainable Interfaces), and finally focusing on the interpretable algorithms at the most internal level (Interpretable Algorithms). The framework suggests iterative cycles of design and evaluation, enabling comprehensive consideration of both algorithmic and human-centred aspects.

In the case of our framework for XAI systems in health and well-being, the process begins with an existing AI system and a set of XAI methods previously applied for verification purposes. For this reason, our focus is on the Evaluation Pole, with the goal is of assessing the AI system and provide insights to improve explainability tasks within the Design Pole.

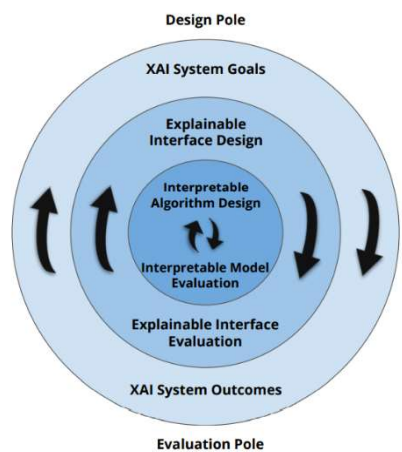

Figure 1: Generic XAI Design and Evaluation Framework (extracted from (Mohseni, Zarei, and Ragan 2021))

## 3 XAI EVALUATION MEASURES

We propose to classify the different explanation qualities three main categories: Machine-Centred Features, Human-Centred Features, and Social-Centred Issues. These categories address different key aspects for assessing XAI systems.

### 3.1 Machine-Centred Features

Machine-Centred features focus on exclusively algorithmic aspects, independent of external evaluators.

**Fidelity**, defined as the correctness of a method in generating true explanations for model predictions (Mohseni, Zarei, and Ragan 2021), is the most extensively studied feature in this category. Fidelity evaluation methods can be divided into:

- **Synthetic Attribution Benchmarks (SABs):** These consist of datasets with ground truth

explanations, created under controlled scenarios. SABs help to identify incorrect methods but cannot confirm their correctness. Various methodologies exist for generating these datasets, including those proposed by (Arias-Duart et al. 2022; Arras et al. 2017; Cortez and Embrechts 2013; Guidotti 2021; Mamalakis, Barnes, and Ebert-Uphoff 2022; Miró-Nicolau, Jaume-i-Capó, and Moyà-Alcover 2024a).

- **Post-hoc Fidelity Metrics**: These metrics approximate fidelity in real-world scenarios where a ground truth explanation is absent. Several authors, including (Alvarez Melis and Jaakkola 2018; Bach et al. 2015; Samek et al. 2017; Rieger and Hansen 2020; Yeh et al. 2019), have proposed post-hoc fidelity metrics. However, these metrics have been criticized for unreliable results (Hedström et al. 2023; Miró-Nicolau, Jaume-i-Capó, and Moyà-Alcover 2025; Tomsett et al. 2020).

**Robustness** is defined as the expectation that minor changes in input data yield similar explanations (Alvarez Melis and Jaakkola 2018). Robustness metrics have been proposed by (Agarwal et al. 2022; Alvarez Melis and Jaakkola 2018; Dasgupta, Frost, and Moshkovitz 2022; Montavon, Samek, and Müller 2018; Yeh et al. 2019).

**Complexity** refers to the amount of variables used in an explanation. Its complementary feature, **sparsity**, ensures that only the truly predictive features contribute to the explanation (Chalasani et al. 2020).

**Localisation** test whether the explanation is focused on a specific region of interest. (Hedström et al. 2023).

**Randomisation** assesses how explanations degrade when data labels or model parameters are randomised, as explored by (Hedström et al. 2023).

## 3.2 Human-Centred Features

Human-Centred Features focus on subjective elements dependent on user interaction with XAI systems. These features, studied beyond AI, have been analysed in social and behavioural sciences (Hoffman et al. 2018; Miller 2019). Key features include **mental models**, **curiosity**, **reliability**, and **trust**, with trust being central to evaluating XAI systems (Barredo Arrieta et al. 2020; Miller 2019). Trust is critical in automation (Adams et al. 2003; Lee and See 2004; Mercado et al. 2016) and is often measured through various scales. (Jian, Bisantz, and Drury 2000) propose measuring trust through an 11-items scale, which has become a de facto standard due to its wide use and influence on other scales.

User Trust is defined as "the attitude that an agent will help achieve an individual's goals in a situation characterized by uncertainty and vulnerability" (Lee and See 2004). While trust is inherently subjective, measuring it objectively is desirable. (Mohseni, Zarei, and Ragan 2021) identify scales and interviews as subjective methods, while (Scharowski et al. 2022) advocate for behavioural measures. (Lai and Tan 2019) propose measuring trust by the frequency of user reliance on system predictions, finding that users trust correct predictions more. Furthermore, (Lai and Tan 2019; Miró-Nicolau et al. 2024) introduce a trust measure that integrates performance and trust data using a confusion matrix, resulting in four distinct measures. These measures, inspired by the well-established True Positive, True Negative, False Positive, and False Negative metrics from classification tasks, provide insights into the interplay between system performance and user trust, allowing for more complex measures, such as Precision and Recall.

## 3.3 Legal and Ethical Issues

Legal and ethical considerations are essential in ensuring transparency and traceability of data and operations within intelligent systems, especially for compliance with regulations and providing legal certainty. In the context of health-related data, it is crucial to address potential biases in algorithms.

According to the General Data Protection Regulation (GDPR) (Union 2016), personal data refers to any information that identifies or can identify a natural person, making privacy a key concern. The Spanish Data Protection Agency highlights the importance of identifying personal data processing, profiling, or decision making related to individuals, which mandates compliance to data protection laws.

While AI systems can use anonymous data, transparency and explicability of algorithms remain essential. Article 78 of the GDPR stipulates that developers should ensure compliance with data protection when designing, selecting, or using applications that process personal data. The principles of **privacy by design** and **privacy by default** (Article 25 of the GDPR) must be prioritized. Additionally, third parties, such as medical personnel, must understand the IA system, its algorithms, and outputs to prevent harm (Justa 2022). These requirements have significant ethical and legal implications, particularly regarding liability.

In 2024, the European Commission approved the AI act, regulating AI technologies within Europe ('Regulation (EU) 2024/1689 Laying down Harmonised Rules on Artificial Intelligence and Amending Regulations (EC) No 300/2008, (EU) No 167/2013, (EU) No 168/2013, (EU) 2018/858, (EU) 2018/1139 and (EU) 2019/2144 and Directives 2014/90/EU, (EU) 2016/797 and (EU) 2020/1828 (Artificial Intelligence Act) – European Sources Online' 2024). This law introduces a risk-based classification system, categorizing AI applications, as Unacceptable, High, Limited, or Minimal risk. Applications in the unacceptable risk category are prohibited, while high-risk and limited-risk applications must meet specific requirements, including human oversight and robustness. Thus, the need for transparency and reliable XAI remains paramount.

Furthermore, the European Commission guidelines for Trustworthy AI ('Assessment List for Trustworthy Artificial Intelligence (ALTAI) for Self-Assessment | Shaping Europe's Digital Future' 2020) define seven requirements for reliable AI: (1) human agency and oversight; (2) technical robustness and safety; (3) privacy and data governance; (4) transparency; (5) diversity, non-discrimination and fairness; (6) societal and environmental well-being; and (7) accountability. These criteria must be evaluated throughout the AI system's lifecycle to ensure legal and ethical compliance.

## 4 XAIHEALTH

In the previous section, we analysed various metrics for evaluating XAI systems, highlighting the significant diversity of approaches used to assess different aspects of explainability. We classified these approaches into three categories: machine-centred, human-centred, and social-centred features. However, it's important to note that these evaluation aspects cannot be assessed simultaneously due to their interdependencies. For instance, if an explanation lacks fidelity to the underlying causes of an AI prediction, the user's trust in that explanation becomes irrelevant, as the prediction itself may be incorrect. Therefore, a comprehensive evaluation of XAI systems must be conducted sequentially.

The framework proposed by (Mohseni, Zarei, and Ragan 2021) requires adaptation for specific contexts, as noted in the introduction. In this section, we present a tailored adaptation of this framework specifically for the healthcare domain. Given that AI systems in healthcare are classified as high-risk under the EU AI Act, they necessitate rigorous verification and monitoring. Our adaptation addresses the unique requirements and challenges associated with integrating XAI into these high-stakes environments.

To facilitate this, we propose a new evaluation framework, named XAIHealth, designed to effectively assess XAI systems within the context of health and well-being. Figure 2 illustrates the adaptation in relation to the foundational framework.

As shown in Figure 2, XAIHealth centres its approach on the Evaluation Pole tasks defined in the base framework (see Figure 1). Each layer, moving from the innermost to the outermost level, comprises two phases: machine-centred analysis and human-centred assessment. These phases are preceded by an initial pre-evaluation phase, which includes training the AI model and applying a XAI method. Legal and ethical considerations are cross-cutting elements that must be addressed throughout the development, evaluation and deployment of the system. Figure 3 illustrates the phases and process flow of the XAIHealth framework.

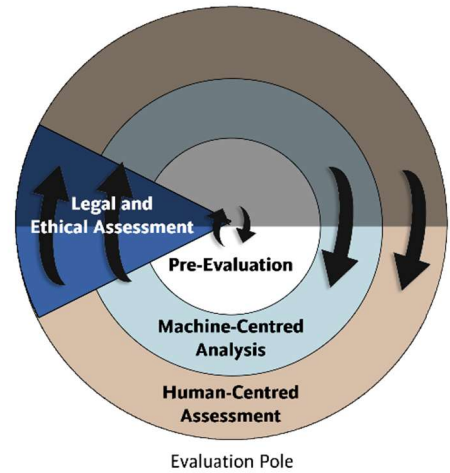

Figure 2: XAIHealth phases in the evaluation pole

### 4.1 Legal and Ethical Assessment

In our framework, ethical and legal factors are integrated into every phase, emphasizing the importance of privacy, data management, and transparency for reliable AI systems in alignment with current legislation.

For privacy and data management, the most restrictive regulations should be prioritized, with European legislation (notable the GDPR) as a reference point due to its rigorous standards. According to Article 4.2 of the GDPR, both, profiling and decision-making about individuals are considered forms of data processing. Health data, categorized as special personal data under Article 9 of the GDPR, requires additional protections. Either anonymous data should be used, or a legal basis must be established for processing. In our case, consent from affected individuals is a viable approach for ensuring compliance.

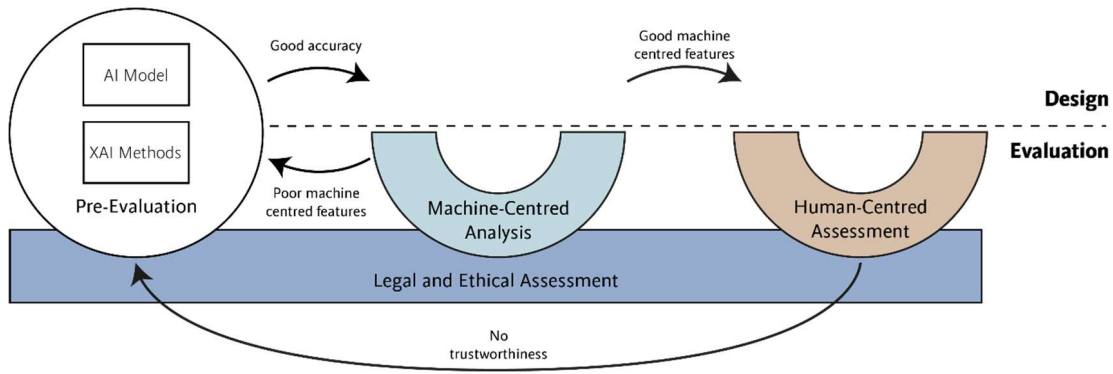

Figure 3: XAIHealth phases and process flow

Transparency is addressed in Article 78 of the GDPR, which mandates that participants must be informed if AI systems will be used in decision-making processes that affect them.

To assess both legal compliance and ethical standards, we propose using the ALTAI (Assessment List for Trustworthy Artificial Intelligence) guidelines ('Assessment List for Trustworthy Artificial Intelligence (ALTAI) for Self-Assessment | Shaping Europe's Digital Future' 2020), developed by an expert group commissioned by the European Commission (AI HLEG). These guidelines operationalize the Ethics Guidelines for Trustworthy AI and align with the principles of the EU AI act.

The ALTAI document identifies seven essential requirements for building trustworthy AI systems. Each requirement entails specific evaluation criteria based on the context. Below, we summarize these requirements:

(1) **Human agency and Oversight**. This ensures that AI systems respect user autonomy and informs users of AI involvement. Six key questions guide the evaluation of whether user autonomy is preserved and whether users are aware they are interacting with AI.
(2) **Technical Robustness and Safety**. This requirement assesses the AI system's resilience to adversarial inputs, novel data, and cybersecurity risks. It focuses on robustness, particularly the system's ability to maintain performance against adversarial alterations (Goodfellow, Shlens, and Szegedy 2015), and varying input data.
(3) **Privacy and Data Governance**. Focused on DGPR compliance, this requirement emphasizes the proper handling and protection of data. Its primary goal is to ensure that data governance practices align with legal privacy standards.
(4) **Transparency**. Transparency refers to how well AI system's processes can be understood. ALTAI specifies that transparency combines explainability with effective communication about the AI system's functions and limitations.
(5) **Diversity**, **Non-discrimination and Fairness**. This requirement addresses the minimization of biases and the promotion of inclusive design. It seeks to eliminate discriminatory elements by ensuring accessibility and universal design principles are considered.
(6) **Societal and Environmental Well-being**. This requirement evaluates the AI system's broader impact, including environmental sustainability, effects on employment and skills, societal influence, and democratic processes. Evaluations are task-dependent and aim to mitigate potential societal risk.
(7) **Accountability**. This requirement focuses on establishing mechanisms for risk management and accountability throughout the system's lifecycle. It enables ongoing monitoring to detect potential errors and risk-prone behaviours.

Each of these requirements aligns with the phases of our proposed framework. As we present each phase, we will outline the relevant ALTAI requirement and discuss strategies for ensuring that the AI system meets these standards. This approach aims to integrate compliance with ethical and legal guidelines into each step of the framework.

## 4.2 Phase 0 – Pre-Evaluation

The Pre-Evaluation phase begins with an existing AI system that has demonstrated adequate efficacy. A set of XAI methods is selected and applied to this system for verification/assessment purposes. The goal is to evaluate holistically the XAI system.

During this phase, three key requirements from the ALTAI guidelines must be addressed: **Privacy and Data governance**, **Diversity, Non-discrimination and Fairness**, and **Societal and Environmental Well-Being**. We will illustrate how each of these requirements is applied in our case study in the following subsection.

### 4.2.1 Application in Case Study

In the Pre-Evaluation phase, we assess an AI system designed to identify signs of pneumonia in chest x-ray images. Specifically, we utilized the AI model proposed in Miró-Nicolau et al. (Miró-Nicolau, Jaume-i-Capó, and Moyà-Alcover 2024a), which proposed a new approach to measure trust in AI systems within a healthcare context. With this experimental setup, the authors trained a well-known Convolutional Neural Network (CNN), ResNet18, introduced by (He et al. 2016). This AI system was trained using a supervised learning approach on a labelled dataset, which includes inputs paired with the correct outputs. The model was trained on 2048 x-ray images from *Hospital Universitari Son Espases* (HUSE*)*, featuring cases of COVID-19 pneumonia and non-pneumonia cases. The system achieved an accuracy of 0.8, a measure of the ratio of correct predictions of total samples. To provide explanations, the GradCAM algorithm (Alvarez Melis and Jaakkola 2018) was applied to produce heatmaps showing which areas of the images were most influential in the model's predictions, giving insight into the model's decision-making process.

From a legal and ethical perspective, the following three requirements from ALTAI must be addressed in this phase:

- **Privacy and Data Governance**. Compliance with GDPR is a primary goal. In our case study, the Research Commission from *Hospital Universitari Son Esp*ases (HUSE) verified data compliance, ensuring all data was anonymized. Thus, we confirm that the Privacy and Data Governance requirement is satisfied.
- **Diversity, Non-Discrimination and Fairness**. This requirement, closely related to privacy and data governance, focuses on ensuring than anonymized x-ray images are free from bias. Accessibility and universal design were also considered in the GUI design process, even so these elements fall outside the scope of the evaluation framework. We conclude that the goal of avoiding discrimination has been met in this case.
- **Societal and Environmental Well-Being**. The model's purpose is to identify COVID-19 pneumonia in patients, and in this context, three main factors must be evaluated: its environmental impact, implications for democracy, and broader societal influence. Firstly, it is clear that this application does not have any direct impact on democratic processes. Similarly, its effect on society is limited, as the system is designed as a diagnostic support tool to enhance medical practice rather than to change social structures.

In terms of environmental impact, the primary concern lies in the energy consumption involved, particularly during the models' training phase, which is generally the most resource-intensive aspect of the process. However, in this case, energy use has been minimized due to two factors: the adoption of a relatively small model (ResNet18) and the use of a modest dataset (2048 images). Thus, we assert that the model meets requirements for environmental sustainability.

## 4.3 Phase 1 – Machine-Centred Analysis

The primary objective of this phase is to evaluate the machine-centred features of the XAI system, focusing on algorithmic attributes that can be measured independently of the end-user. Fidelity is considered the most crucial metric in this category, as shown by its predominance in the state-of-the-art. However, calculating fidelity is challenging due, as (Hedström et al. 2023) note "since the evaluation function is applied to the results of the unverifiable explanation function, the evaluation outcome also renders unverifiable". To address this, we consider necessary to adopt a validated post-hoc measurement approach, using only validated metrics to avoid unreliable evaluations.

In our analysis, we excluded unvalidated metrics, therefore the lack of verification in existing post-hoc fidelity metrics makes them unsuitable for real-world applications. (Hedström et al. 2023) proposed a meta-evaluation process that reviewed several machine-centred metrics, revealing both strengths and weaknesses. However, none of the ten metrics analysed achieved perfect results, meaning fidelity and similar features may not yet be reliably usable in real-world contexts.

Among the machine-centred features, robustness has also been largely studied. Miró-Nicolau et al. (Miró-Nicolau, Jaume-i-Capó, and Moyà-Alcover 2024b), developed a set of tests for assessing robustness metrics, initially applying them to the AVG-Sensitivity and MAX-Sensitivity by (Yeh et al. 2019). We extended these tests to include other robustness metrics, identifying Local Lipschitz Estimate (LLE) by (Alvarez Melis and Jaakkola 2018) as a reliable and practical option, being . LLE is the only metric that passed both robustness tests, making it our primary criterion for assessing explanation robustness. However, if additional metrics are developed and validates in the future, they can be incorporated into the framework without requiring further modifications.

This phase also addresses two ALTAI requirements: Technical Robustness and Safety and Transparency. If issues arise during this phase, it may be necessary to revisit the Pre-evaluation phase to determine whether the shortcomings stem from the AI model or the XAI method itself.

### 4.3.1 Application in Case Study

To evaluate the robustness of explanations in our case study, we applied the Local Lipschitz Estimate (LLE) proposed by (Alvarez Melis and Jaakkola 2018) to GradCAM generated explanations. The optimal LLE score is 0, representing maximum robustness, while 1 is the worst possible outcome. We established a flexible acceptance threshold, considering an explanation robust if its results fall within the top 10%. This was because the fact that even a perfect XAI algorithm may exhibit slight robustness limitations due to the underlying AI model itself, making a minor margin of error acceptable. We obtained a mean LLE value of 0,082 with a standard deviation of 0,108. This results show that the system's robustness falls within the accepted threshold, fulfilling the Technical Robustness and Safety requirement of the ALTAI guidelines. Additionally, Transparency – a core goal of any XAI approach and a key component in mitigating black-box limitations – is addressed by the GradCAM method, which offers an interpretable heatmap-based explanation of model predictions. These results confirm that the system is ready to proceed to the next evaluation phase.

## 4.4 Phase 2 - Human-Centred Assessment

The purpose of this phase is to evaluate human-centred features of the XAI method which are directly influenced by the end-user's experience. To account for this, an interface that displays both the AI system's prediction and the accompanying explanation form the XAI method must be utilized.

This phase encompasses several aspects, with trust being a primary focus in the XAI field, as highlighted by Miller (Miller 2019). Trust can be assessed from two perspectives: as an attitude (the user's self-perception) or as a behaviour (following the AI system's advice), according to (Scharowski et al. 2022). These two approaches are clearly related to different evaluation methods: objective (behavioural) or subjective (attitudinal).

To measure trust effectively, we adopt pre-existing metrics reviewed in the previous section. We recommend the metric proposed by (Miró-Nicolau et al. 2024), which uses a behavioural approach that incorporates the AI system's prediction performance into the trust evaluation.

If the outcomes of this phase indicate issues, it may be necessary to revisit the pre-evaluation phase to identify whether issues stem from the interface design or broader XAI system limitations.

In this phase, the ALTAI requirement of Human Agency and Oversight must be addressed, primarily by ensuring users are informed at all times that they are interacting with an AI system. This requirement closely aligns with the focus of this phase, which emphasizes user interface design and clear information communications between the user and the AI system.

### 4.4.1 Application in Case Study

In our case study, which used x-ray images, we assessed whether radiologists and other specialists trusted the AI system's outputs. Results of this phase, along with the trust measurement, were published by Miró-Nicolau et al. (Miró-Nicolau et al. 2024). The design team developed an interface that presented the AI system's prediction and the XAI method-generated explanation simultaneously to the end-users. To simplify understanding, the explanations were refined by removing less significant pixels using various threshold values. Specialists then rated their level of trust in the combined diagnosis and explanation, with their responses assessed using (Miró-Nicolau et al. 2024) trust metric.

The Human Agency and Oversight ALTAI requirement is fully consistent with the interface design in this case study. Specifically, the interface clearly communicates that the AI model and XAI method are in use, fulfilling this ALTAI requirement.

The outcomes of this phase are summarized in Table 2. The table shows trust metrics combining classification accuracy with user trust. A complete trust in the model would be reflected by a value of 1 across all three metrics. However, the results reveal that users did not fully trust the model, prompting a return to Phase 0 to improve the system's reliability. Thus, until trust is restored, this system is not suitable for real-world deployment.

Table 2: Trust results from (Miró-Nicolau et al. 2024)

| Metric | User 1 | User 2 | Mean |
| --- | --- | --- | --- |
| Precision | 0.1094 | 0.0156 | 0.0625 |
| Recall | 0.3333 | 0.0714 | 0.2022 |
| F1-Score | 0.1647 | 0.0256 | 0.0952 |

Despite the lower trust levels observed, these results demonstrate the effectiveness of our framework in identifying limitations within AI systems, ensuring that only trustworthy models proceed to real-world application.

### 4.5 XAI System Operation

It is essential to monitor health tools during public use to assess their long-term effects. Consequently, implementing a monitoring phase is critical to ensure that the XAI system can continuously provide services in compliance with the specified requirements. This need aligns with the seven requirements outlined in the ALTAI guidelines. Current legislation, specifically the EU AI Act, mandates that high-risk AI deployments, such as those related to health and well-being, undergo appropriate risk assessment and mitigation strategies. Furthermore, XAI systems in healthcare must be subject to review, approval, and ongoing monitoring by an institutional ethics committee. If necessary, corrections, modifications, and enhancements should be made to the deployed system.

## 5 CONCLUSIONS

In this paper, we present **XAIHealth**, a new evaluation framework specifically designed for XAI Systems for health and well-being. The framework is built by taking the general guidelines for a multidisciplinary approach to develop XAI systems, as detailed in (Mohseni, Zarei, and Ragan 2021).

The result is a structured evaluation framework that comprises two main phases: **Machine-Centred Analysis**, and **Human-Centred Assessment**. These phases are preceded by a **Pre-Evaluation** stage and conclude with the **XAI System Op**eration phase. Iteration and feedback are integral throughout the framework's phases, and legal and ethical considerations are addressed at every step of the evaluation process.

While our proposal primarily targets health and well-being applications, we believe that this framework could also be applicable to any XAI system that significantly influences human behaviour.

Future work will focus on two key areas. First, we will elaborate on the tasks necessary during the system operation phase, specifically concerning monitoring and improvement. Second, we aim to apply the framework to additional case studies to validate its effectiveness and identify potential enhancements.

## ACKNOWLEDGEMENTS

This work is part of the Project PID2023-149079OB-I00 funded by MICIU/AEI/10.13039/501100011033 and by ERDF/EU.